\pdfoutput=1

\documentclass[11pt]{article}

\usepackage[preprint]{acl}
\usepackage{times}
\usepackage{latexsym}
\usepackage[T1]{fontenc}
\usepackage[utf8]{inputenc}

\usepackage{amsfonts}
\usepackage{amsmath}

\usepackage{float}

\usepackage{mathtools}
\usepackage{twemojis}

\usepackage{cellspace}
\newcommand\cincludegraphics[2][]{\raisebox{-0.4\height}{\includegraphics[#1]{#2}}}
\usepackage{setspace}

\DeclarePairedDelimiter\abs{\lvert}{\rvert}%
\DeclarePairedDelimiter\norm{\lVert}{\rVert}%

\makeatletter
\let\oldabs\abs
\def\abs{\@ifstar{\oldabs}{\oldabs*}}
\let\oldnorm\norm
\def\norm{\@ifstar{\oldnorm}{\oldnorm*}}
\makeatother

\usepackage{microtype}

\usepackage{inconsolata}

\usepackage{graphicx}

\usepackage[group-separator={,},group-digits=integer]{siunitx}
\usepackage{booktabs}

\usepackage{enumitem}

\title{Investigating ReLoRA: Effects on the Learning Dynamics of Small~Language~Models}

\author{
    \textbf{Yuval Weiss}\thanks{Corresponding author: \texttt{yw580@cantab.ac.uk}} \qquad
    \textbf{David Demitri Africa} \\
    \textbf{Paula Buttery} \qquad
    \textbf{Richard Diehl Martinez} \\
    University of Cambridge
}

\begin{document}
\maketitle
\begin{abstract}
Parameter-efficient methods like LoRA have revolutionised large language model (LLM) fine-tuning. ReLoRA extends this idea to pretraining by repeatedly merging and reinitialising low-rank adapters, increasing cumulative rank while keeping updates cheap. This aligns well with observations that high-capacity models learn through locally low-rank trajectories that expand over time. By contrast, recent work suggests that small language models (SLMs) exhibit rank deficiencies and under-utilise their available dimensionality. This raises a natural question: can ReLoRA's rank-expanding update rule \textit{steer} SLMs toward healthier learning dynamics, mitigating rank bottlenecks in a capacity-constrained regime? We argue SLMs are an ideal testbed: they train quickly, enable controlled ablations, and make rank phenomena more measurable. 
We present the first systematic study of ReLoRA in SLMs (11M–66M parameters), evaluating both performance and learning dynamics. Across loss, Paloma perplexity, and BLiMP, we find that ReLoRA underperforms full-rank training, with gaps widening at larger scales. Analysis of proportional effective rank and condition numbers shows that ReLoRA amplifies existing rank deficiencies and induces ill-conditioned updates early in training. Our results suggest that while ReLoRA's merge-and-restart strategy can expand ranks in larger models, it does not straightforwardly translate to capacity-limited SLMs, motivating adaptive-rank or hybrid-rank approaches for low-compute pretraining.

\end{abstract}

\begin{tabular}{@{}c l@{}}
  \cincludegraphics[width=1.9em]{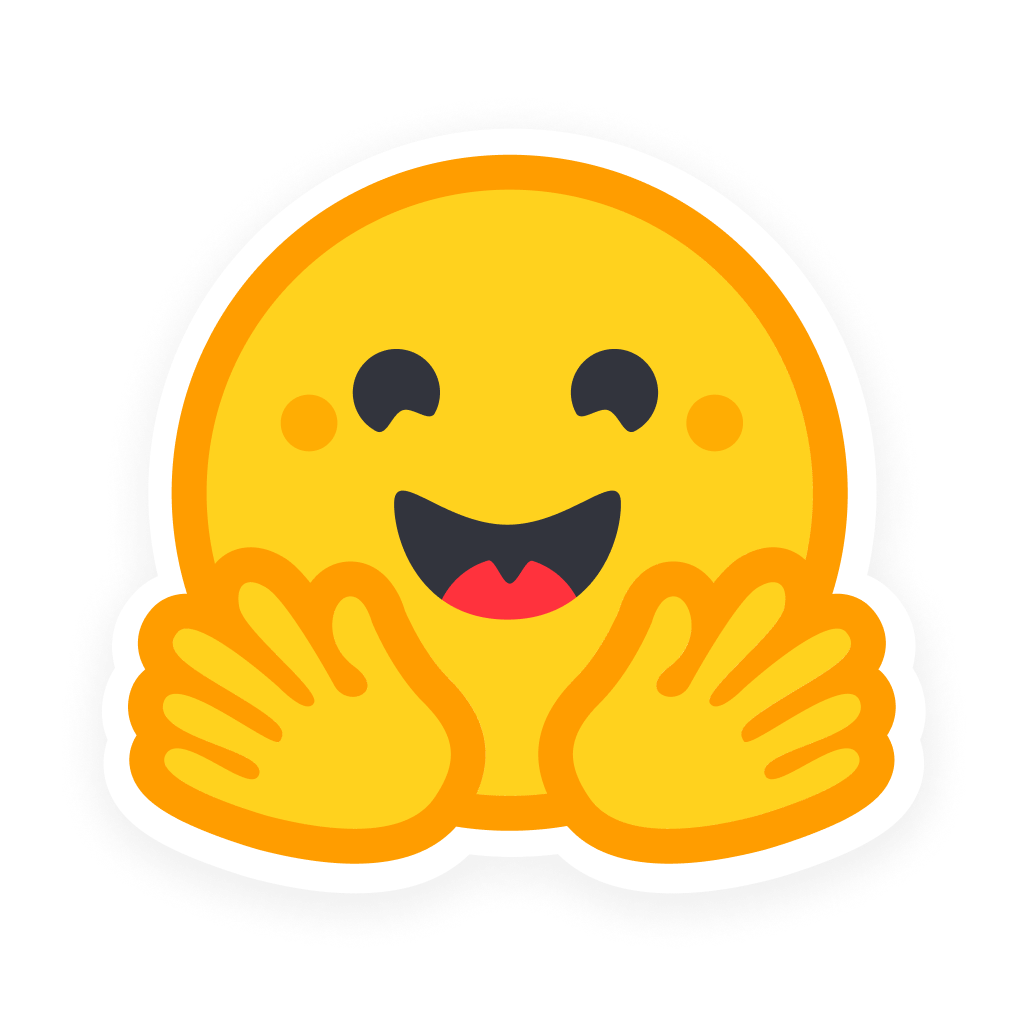} &
  \footnotesize{\texttt{\href{https://huggingface.co/collections/yuvalw/pico-relora-6766e4e726eb8811626915c0}{yuvalw/pico-relora}}} \\
  \cincludegraphics[width=1.5em]{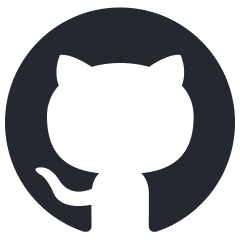} &
  \footnotesize{\texttt{\href{https://github.com/Yu-val-weiss/pico-relora}{Yu-val-weiss/pico-relora}}}
\end{tabular}

\section{Introduction}


Contemporary work on language modelling has continually prioritised ever-larger scales, delivering remarkable capability gains \citep{chowdhery_palm_2023, grattafiori_llama_2024, openai_gpt-4o_2024}. However, the computational and environmental costs of large-scale language modelling are substantial \citep{chowdhery_palm_2023, grattafiori_llama_2024, morrison_holistically_2024, openai_gpt-4o_2024}. Small language models (SLMs) offer a complementary path: they are cheap to train and deploy, easier to study, and attractive for settings where safety, privacy, or energy constraints dominate \citep{schwartz_green_2020}. However, SLMs lag in quality due to tight capacity limits and brittle optimisation \citep{mielke_what_2019, ettinger_what_2020, kaplan_scaling_2020}.


\paragraph{A rank-centric view of learning.} A growing body of work frames transformer learning through the lens of the rank of model weights. In high-capacity models, training often proceeds via low-rank updates that \textit{gradually} expand effective rank \citep{aghajanyan_intrinsic_2021, boix-adsera_transformers_2023}. Conversely, SLMs show rank deficiencies and anisotropic representations that restrict gradient flow and under-utilise model dimensionality \citep{noci_signal_2022, diehl_martinez_tending_2024, godey_anisotropy_2024}. From this perspective, methods that shape the rank profile of updates could directly influence optimisation quality and final capability.

\paragraph{LoRA.}
LoRA \citep{hu_lora_2022} has been widely used due to its effectiveness and applicability to any model with matrix computations, enabling straightforward fine-tuning of even the largest models \citep{blattmann_stable_2023,dettmers_qlora_2023,fomenko_note_2024}. LoRA introduces low-rank adaptation, where the pretrained model weights are frozen, and fine-tuning is performed with trainable rank decomposition matrices at each layer of the architecture, thus significantly reducing the number of trainable parameters. 

\paragraph{ReLoRA.}
ReLoRA, proposed by \cite{lialin_relora_2023}, aims to leverage LoRA's overwhelming success in low-rank adaptation for fine-tuning to advance efficient pretraining. It does this by injecting low-rank LoRA-style matrices into a language model, then repeatedly merging and reinitialising them throughout the training process, thereby parametrising gradient updates through the merge operations at each restart. The method is tested on models with 60M to 1.3B parameters, improving training speeds and reducing GPU memory footprints, especially at the higher end. However, research for models even smaller than this is limited. 

\paragraph{Why ReLoRA might help SLMs.} Small transformer models suffer from rank deficiencies in their weight matrices, limiting the subspace they can explore during training \citep{diehl_martinez_tending_2024}. ReLoRA \citep{lialin_relora_2023} periodically merges low-rank adapters into the base weights and reinitialises them. Summing distinct low-rank updates increases the rank of the \textit{cumulative} update, potentially widening the subspace explored over training. If SLMs struggle because their updates remain trapped in low-rank bottlenecks, then an explicit rank-expansion mechanism could \textit{steer} them toward healthier dynamics, improving sample efficiency and language competence in the low compute regime. This provides the intuition behind our methodology. 

\paragraph{Why test this in SLMs?}
SLMs are fast to train, enabling careful ablations across model structure alongside detailed diagnostics such as effective rank and condition number that would otherwise be prohibitively costly in billion-parameter models. Their capacity limits also amplify rank phenomena, making both benefits and failure modes easier to detect. Therefore, SLMs are a sensitive and economical \textit{sandbox} for understanding how rank-structured updates shape learning.

\paragraph{This work: \textit{research question}.}
In this work, we ask whether ReLoRA's rank-expanding merge-and-restart methodology actually helps in the capacity-constrained regime. In other words, does ReLoRA \textbf{boost} performance or \textbf{drag} it down? The case for either can be summarised as follows:

\texttwemoji{thumbsup} \textbf{Boost}: By strategically resetting and merging LoRA matrices, ReLoRA aggregates multiple low-rank steps, which may widen those bottlenecks and enable smaller models to capture more complex patterns

\texttwemoji{thumbsdown} \textbf{Drag}: However, ReLoRA may also have the opposite effect: further reducing the (already limited) representational capacity of SLMs.

\paragraph{This work: \textit{outline of methodology and results}.}
We extend a Llama-style SLM \citep{touvron_llama_2023, diehl_martinez_pico_2025} with ReLoRA and run matched ablations at 11M and 66M parameters. We evaluate training loss, Paloma perplexity \citep{magnusson_paloma_2024}, and BLiMP \citep{warstadt_blimp_2020}, and additionally probe learning dynamics via proportional effective rank (PER) and condition numbers of both weights and updates. Contrary to the motivating intuition, ReLoRA \textit{does not} boost SLMs: it underperforms full-rank training and reinforces rank deficiencies, with early training marked by highly ill-conditioned updates. We thus conclude that the mechanism that helps larger models does not straightforwardly transfer to SLMs.

\paragraph{This work: \textit{contributions}.}
Our contributions are as follows:

\begin{enumerate}[label=\textbf{(\arabic*)}]
    \item \textbf{Novel systematic evaluation of ReLoRA for small language models (SLMs).} We investigate ReLoRA on 11M and 66M parameter models, filling a critical gap in understanding its behaviour for low-compute domains. We make two public code contributions: a public HuggingFace space
    to be used for evaluating LMs on the BLiMP task through the \texttt{evaluate} library and a public fork, \texttt{pico-relora}, 
    of \texttt{pico-train} \citep{diehl_martinez_pico_2025} extending \texttt{pico-decoder} with ReLoRA.

    \item \textbf{Comprehensive analysis of learning dynamics.} We compute proportional effective rank (PER) and condition numbers of weights and gradient updates throughout training, revealing that ReLoRA consistently reduces rank and induces highly ill-conditioned updates in SLMs.

    \item \textbf{\texttwemoji{thumbsdown} Drag: Evidence of performance degradation and training instability.} Across loss, Paloma perplexity, and BLiMP evaluation, ReLoRA underperforms its full-rank baseline, with widening gaps as model size increases from the \texttt{tiny} to \texttt{small} scales. Our findings suggest that low-rank strategies do not trivially transfer to SLMs and thus motivate hybrid or adaptive-rank approaches for future low-resource model training.
\end{enumerate}

\begin{figure*}
    \centering
    \includegraphics[width=\linewidth]{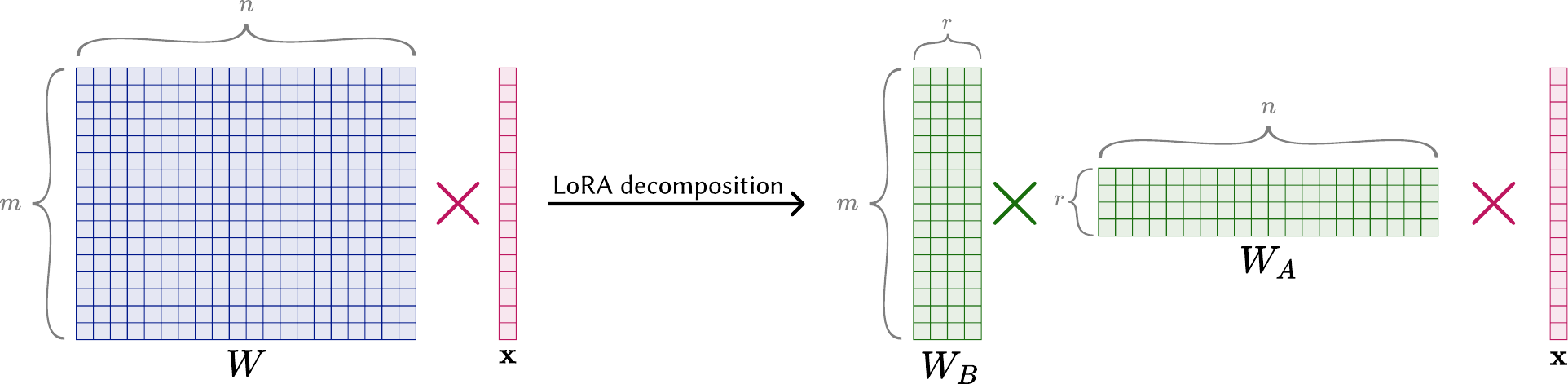}
    \caption{LoRA decomposition \citep{hu_lora_2022}, which can be applied to any linear operation parameterised by a matrix  }
    \label{fig:lora-dec}
\end{figure*}

\section{Background: ReLoRA}

We now introduce ReLoRA, a method utilising low-rank updates to train high-rank neural networks more efficiently \citep{lialin_relora_2023}.

\paragraph{Motivation: LoRA to ReLoRA.}
ReLoRA is based on LoRA, a vastly successful parameter-efficient fine-tuning technique which leverages `LoRA decomposition' matrices to effectively fine-tune even the very largest models \citep{hu_lora_2022, fomenko_note_2024}.  Given LoRA's success, \cite{lialin_relora_2023} propose ReLoRA, which applies the same decomposition methodology to the \textit{pretraining} of an LM, providing competitive performance for significantly less compute. 


\paragraph{LoRA decomposition.} In a neural network, a linear operation parameterised by $ W\in\mathbb R^{m \times n}$ applied to a token $\mathbf{x} \in \mathbb R^{n}$ has a gradient denoted $\Delta W$. After a gradient update, the output of the linear operation is changed by $(W + \Delta W)x =Wx + \Delta W x$ . LoRA proposes decomposing the gradient update $\Delta W$ to two matrices $W_A \in \mathbb R^{r\times n}$ and $W_B \in \mathbb R^{m \times r}$, for some very small LoRA parameter $r \ll \min(m, n)$ and scaling parameter $s$, as shown in Equation~\ref{eq:lora-dec} and Figure~\ref{fig:lora-dec}. These matrices are known as adapters. 
\begin{equation}\label{eq:lora-dec}
    \Delta W = s W_BW_A
\end{equation}
For small $r$, the number of parameters being directly updated is $r(m + n) < m\cdot n$. 

LoRA is implemented by injecting trainable linear layers parameterised by $W_A$ and $W_B$ into the model. Throughout this work, we refer to $W_A$ and $W_B$ collectively as the LoRA module, and $W$ as the base matrix. 

\paragraph{ReLoRA extends LoRA decomposition by introducing restarts.}
This is motivated by a core property of matrices: for some matrix $A$, there exists a matrix $B$ such that
\begin{equation}
    \mathrm{rank}(A + B) \geq \max \left(\mathrm{rank}(A), \mathrm{rank}(B)\right)
\end{equation}
This means that combining different low-rank updates can result in a higher-rank update overall.

ReLoRA proposes periodically restarting LoRA's low-rank adapters. At each restart, the low-rank matrices $W_A$ and $W_B$ are merged back into $W$ (by adding $W_BW_A$ to $W$), and reinitialised for further training. After N such restarts, the cumulative update $\Delta W_\mathrm{eff}$ is thus
\begin{equation}\label{eq:lora-restart}
    \Delta W_\mathrm{eff}= s \sum_{i=1}^N W_B^iW_A^i
\end{equation}
where each pair of $W_A^i$ and $W_B^i$ comes from a distinct stage of ReLoRA in between each restart, and $s$ is a scaling parameter. This assumes that the LoRA modules have identical rank $r$. The equation only holds for AdamW \citep{loshchilov_decoupled_2019} without weight decay on $W$, thus requiring the optimisation modifications described below. 

This strategy effectively increases the rank of the total update $\Delta W_\mathrm{eff}$ even though individual terms remain low-rank. As a result, ReLoRA can represent more complex updates than LoRA with the same number of trainable parameters per step, achieving higher representational effectiveness while maintaining parameter efficiency. The motivation for this approach is grounded in prior observations that neural network training often operates in a locally low-rank regime \citep{aghajanyan_intrinsic_2021, boix-adsera_transformers_2023}.  Understanding how these low-rank dynamics evolve requires situating ReLoRA within the broader literature on rank growth and intrinsic dimensionality.


\section{Background: Rank Dynamics}

\citet{lialin_relora_2023} measure the singular value spectra produced by models trained with ReLoRA and find that these models resemble models trained with normal full-rank training, particularly when comparing larger singular values. While these results suggest that ReLoRA can retain some of the representational richness of full-rank training, its behaviour in much smaller models remains untested.

This is particularly relevant in light of prior work on intrinsic dimensionality, which measures the minimal number of parameters required to solve a task to a given accuracy. \citet{aghajanyan_intrinsic_2021} show that large pretrained LMs occupy surprisingly low-dimensional subspaces when fine-tuned, sometimes requiring only a few hundred effective parameters. Yet these measurements are limited to a 125M-parameter RoBERTa\textsubscript{BASE} \citep{liu_roberta_2019-2} and focus on fine-tuning rather than pretraining. In the pretraining domain, \citet{diehl_martinez_tending_2024} find that smaller models in general tend to use their available capacity less effectively than larger models. 

Relatedly, studies of rank dynamics in transformers have shown two interesting effects: first, weight matrices tend to increase in effective rank gradually over training \citep{boix-adsera_transformers_2023}, expanding their representational capacity; second, at the same time, “rank collapse” at initialisation can severely limit early gradient flow in key and query projections \citep{noci_signal_2022}. These findings suggest that the ability to build rank over time is important for model quality. How ReLoRA intersects with these findings in the context of small language models is an open question and the subject of this paper. We hypothesise two possible effects.

ReLoRA's repeated low-rank resets could either \textbf{boost} performance and facilitate rank growth by exploring new subspaces, or \textbf{drag} performance by exacerbating rank bottlenecks in small models. Understanding which effect dominates requires probing ReLoRA in precisely these capacity-limited regimes.

\section{Methodology}

The model implementations and training code are forked from \texttt{pico-train} \cite{diehl_martinez_pico_2025}, a lightweight framework for training language models. \texttt{pico-train} implements \texttt{pico-decoder}, a Llama-style decoder \citep{touvron_llama_2023} illustrated and described in Figure~\ref{fig:pico-decoder} in Appendix~\ref{sec:pico-decoder}. \texttt{pico-decoder} is extended with an implementation of ReLoRA \citep{lialin_relora_2023} to form \texttt{pico-relora}. The comparison of these two models forms the basis of the methodology presented in this paper. 

\subsection{Experimental setup} \label{subsec:exp_setup}

The experimental setup is as follows: We execute two \texttt{pico-relora} training runs at \texttt{tiny} and \texttt{small} scales to compare to equivalent (other than the presence of ReLoRA) baseline \texttt{pico-decoder} models. This forms a targeted ablation study, consisting of four runs at two scales, by which the behaviour of ReLoRA in small language models can be systematically analysed. The runs' respective numbers of trainable and total parameters are shown in Table~\ref{tab:num-params}.

\begin{table}[t]
    \centering
    \begin{tabular}{lS[table-format=8]S[table-format=8]}
    \toprule
     \textbf{Model}    & \textbf{Trainable params} & \textbf{Total params} \\
    \midrule 
    \texttt{t-dec} & 11282784 & 11282784 \\
    \texttt{s-dec} & 64595328 & 64595328 \\
    \texttt{t-rel} & 10060128 & 11682144 \\
    \texttt{s-rel} & 40240512 & 66192768 \\
    \bottomrule
    \end{tabular}
    \caption{Trainable and total numbers of parameters for the \texttt{tiny} and \texttt{small} models. \texttt{t} = \texttt{tiny}, \texttt{s} = \texttt{small}, \texttt{dec} = \texttt{decoder}, \texttt{rel} = \texttt{relora}.}
    \label{tab:num-params}
\end{table}

The runs were executed for \num{20000} batch steps each, resulting in each model encountering a total of 41.9B tokens throughout training. Training infrastructure and experiment runtimes are described in Appendix~\ref{sec:infrastructure}. 

Table~\ref{tab:relora_config} depicts the configuration parameters for ReLoRA, with the full configurations shown in Tables~\ref{tab:lr_opt_config}, \ref{tab:data_config} and \ref{tab:model_config} of Appendix~\ref{app:full_conf}. These are identical between the two runs. ReLoRA modules are injected into each linear layer of the attention and feed-forward layers. The modules are set to use a fixed (trainable scaling is set to \texttt{False}) scaling parameter $s = \frac{\alpha}{r} = \frac{32}{16} = 2$, a dropout probability of $0.1$ and an internal LoRA parameter $r$ of $16$. ReLoRA resets are configured to occur every $2000$ optimiser steps. These values are based on the ones in \citet{lialin_relora_2023}. We omit the 20k step full-rank warm-start for a purer analysis of learning dynamics and to avoid the additional computational overhead. 

\begin{table}[t]
    \centering
    \begin{tabular}{lr}
    \toprule
      \textbf{Parameter} & \textbf{Value} \\
    \midrule
       Target modules  & \texttt{attention \& swiglu}  \\
       Reset frequency & $2000$ \\ 
       $r$ & $16$  \\
       Trainable scaling & \texttt{False} \\
       $\alpha$ & $32$ \\
       Dropout & $0.1$ \\ 
    \bottomrule 
    \end{tabular}
    \caption{ReLoRA configuration for training runs. $r$ is the LoRA parameter, and $\alpha$ configures the scaling parameter $s$ according to $s = \frac{\alpha}{r}$}
    \label{tab:relora_config}
\end{table}

\paragraph{ReLoRA restarts require modifications to the optimiser and learning rate scheduler.} 
\texttt{pico-decoder} uses AdamW \citep{loshchilov_decoupled_2019} as its optimiser, which includes moments accumulated over previous gradient update steps in its parameter update calculations. Update step $i+1$ will thus still be guided by gradients calculated in step $i$. Therefore, ReLoRA randomly prunes a (configurable) proportion of the optimiser states (around $99\%$ of them) to zero to prevent continuing in the same trajectory as before the reset, which could hinder learning new subspaces. The learning rate is simultaneously set to zero to avoid divergent losses and linearly `re-warmed' according to a jagged cosine scheduler. The two learning rate schedulers are illustrated in Figure~\ref{fig:lr-schedulers}.

\subsection{Training and evaluation datasets: Dolma, Paloma and BLiMP} \label{sec:impl-datasets}

\paragraph{Training on Dolma, a three-trillion-token English dataset.}
\texttt{pico-train} is set up to use a pre-tokenised, pre-shuffled version of Dolma \citep{soldaini_dolma_2024}, an open-source, three-trillion-token English dataset.\footnote{This version of Dolma is made available on HuggingFace here: \url{https://huggingface.co/datasets/pico-lm/pretokenized-dolma}.}
Dolma is composed of scientific writing, website content, computer code, books in the public domain, social media data and material from encyclopedias. Dolma considers both ethical and legal issues throughout the data curation process, avoiding sources that contain copyrighted materials or personally identifiable information. 

\paragraph{Perplexity is computed on the Paloma benchmark dataset.}
In this work, perplexity is computed on Paloma, a benchmark dataset that measures LM fit over \qty{546} different English and computer code domains \citep{magnusson_paloma_2024}. The goal of Paloma is to avoid the assumption that an LM suited to one domain will generalise well to others. It uses stratified subsampling based on empirical estimates of variance, ensuring domains are equally represented. Paloma contains \qty{123683201} tokens across its test and validation splits; thus, a subsampled version of the corpus, termed \texttt{palomy-tinsy}\footnote{This pre-processed, subsampled dataset is available here: \url{https://huggingface.co/datasets/pico-lm/pretokenized-paloma-tinsy}.}
is used. Like \texttt{pretokenized-dolma}, \texttt{palomy-tinsy} is pre-tokenised and pre-shuffled. It consists of 1.44 thousand data points in a single validation split.

\paragraph{Evaluating linguistic understanding with BLiMP.} 

The Benchmark of Linguistic Minimal Pairs for English (BLiMP) is a `challenge set' for evaluating the understanding of significant grammatical phenomena in English \citep{warstadt_blimp_2020}. BLiMP is composed of 67 distinct datasets, in turn consisting of 1000 minimal pairs, such as `there was \textbf{[\textit{bound} / \textit{unable}]} to be a fish escaping'. For each minimal pair in BLiMP, one sentence is deemed grammatically acceptable, and the other is considered unacceptable. Each set of minimal pairs targets a specific linguistic phenomenon. The 67 phenomena are aggregated further into twelve broader categories. A model evaluated on BLiMP is expected to assign a higher log-likelihood to the acceptable sentence for each minimal pair in the dataset. Its final score is simply the proportion of minimal pairs for which it correctly prefers the acceptable sentence.

\section{Results and evaluation}

\subsection{Main evaluation}

\begin{figure}[t]
    \centering
    \includegraphics[width=\linewidth]{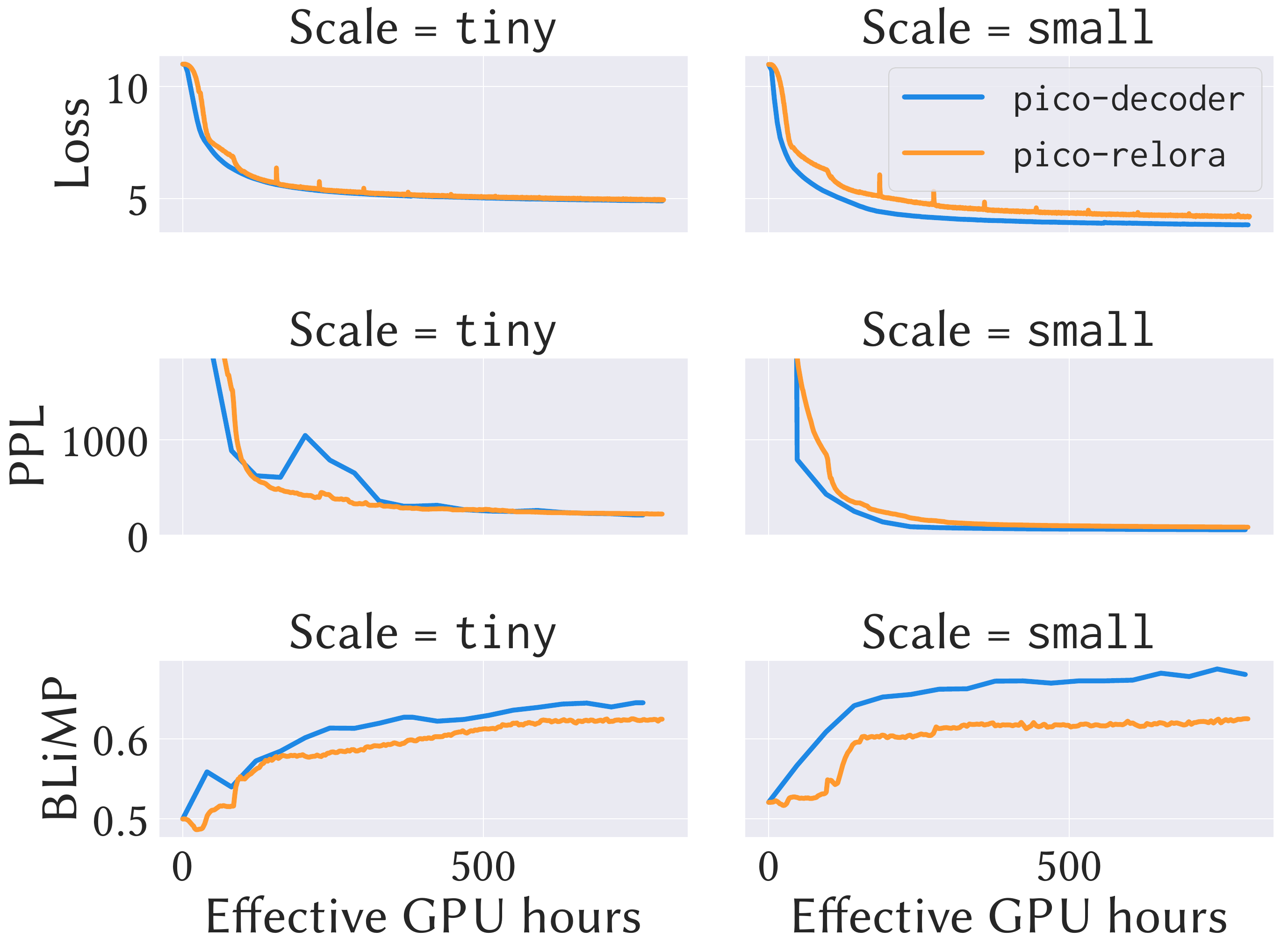}
    \caption{Training trajectories of cross-entropy loss,  Paloma perplexity  and  BLiMP score across the \texttt{tiny} and \texttt{small} models, plotted against GPU hours taken}
    \label{fig:train-traj-gpu}
\end{figure}

This section examines the training trajectories of the four models, providing an overview of their learning processes at a high level. Figure~\ref{fig:train-traj-gpu} depicts each model's cross-entropy loss, Paloma perplexity and overall BLiMP score throughout training, measured against effective GPU hours.

As Figure~\ref{fig:train-traj-gpu} illustrates, \texttt{pico-relora} performs almost identically to \texttt{pico-decoder} on the \texttt{tiny} scale, with a more pronounced difference for \texttt{small}. This difference is larger for loss than for perplexity. The loss values in the \texttt{pico-relora} models show small spikes, coinciding with ReLoRA restarts, which lead to a localised training instability that the models quickly recover from. 
 
\texttt{pico-decoder} deviates from the expected perplexity trajectory in the \texttt{tiny} run, likely due to the model finding a suboptimal local minimum. \texttt{pico-relora} does not experience the same deviation, which may indicate higher training stability, but this can only be confirmed by executing additional training runs. This highlights an important limitation of the results throughout this work: only one training run was executed for each model due to time and computational restrictions. 

Further to the above findings for loss and perplexity, the \texttt{pico-relora} models underperform on BLiMP even more, with pronounced differences at the \texttt{small} scale. These results are significant at the $1 \times 10^{-10}$ level, according to a proportion Z-test set up with an alternative hypothesis that the \texttt{pico-decoder} models perform better. 

\subsection{Analysis}

In this section, given \texttt{pico-relora}'s underachievement, we strive to identify the inherent properties of the model's training dynamics that may have caused the observed performance drag. To do so, we make use of the `residual stream' framework \citep{elhage_mathematical_2021} through analysis introduced in \citet{diehl_martinez_tending_2024}.

\subsubsection{Preparation}

\paragraph{The `residual stream' framework for analysing LMs.} 
The `residual stream' is a mathematical framework for analysing autoregressive decoder-only LMs \citep{elhage_mathematical_2021}, which focuses on `residual connections'. The framework considers residual blocks composed of an attention layer and a feed-forward layer, omitting layer normalisation for simplicity. Each of these layers `reads' from the residual stream with a linear projection and then `writes' back to it using another linear projection.

\paragraph{Splitting attention head terms into `Output-Value circuits'.}
Attention heads can be conceptualised as being composed of two essentially independent calculations: a `Query-Key' circuit, which calculates the attention pattern, and an `Output-Value' (OV) circuit, which computes the amount a given token affects output logits (if it is attended to) \citep{elhage_mathematical_2021}. The OV circuit is calculated by taking the matrix product of the Output and Value matrices. OV circuits, despite being parameterised separately, may be considered as individual, low-rank matrices. We focus on the OV circuit as it `writes' to the residual stream, a part of the model that has been shown to suffer from performance bottlenecks via the output representations of the model \citep{godey_why_2024}.

\paragraph{Learning dynamics \& investigated parameters.}

Following this, learning dynamics analysis is performed on certain activations and parameters of the models' attention and feed-forward layers. Specifically, we choose to analyse activations and parameters that `write' to the residual stream, namely the OV circuit and SwiGLU $W_2$ layers \citep{elhage_mathematical_2021}. Therefore, their behaviour has a salient influence on the evolution of internal representations across layers. This makes them particularly valuable probes for rank bottlenecks and deficiencies. 

\subsubsection{Proportional effective rank}

Given what we know about the rank deficiencies in smaller models, we aim to investigate how they behave and evolve. To achieve this, we utilise PER \citep{diehl_martinez_tending_2024}, a metric that enables size-agnostic comparison of the effective rank of parameter weight matrices. We measure PER on the weight matrices in the feed-forward and attention layers that parameterise the `writes' back to the residual stream. PER is given by Equation~\ref{eq:per} below.

Let the parameter at layer $l$ be $\theta_l \in \mathbb{R}^{d_\mathrm{model} \times d_\mathrm{inter}}$ where $d_\mathrm{inter}$ is the dimension of the intermediate representations in either the attention or feed-forward layers. 
\begin{equation}\label{eq:per}
    \mathrm{PER}(\theta_l) = \frac{\mathrm{ER}(\theta_l)}{d_\mathrm{inter}}
\end{equation}
where $\mathrm{ER}$ \citep{roy_effective_2007}  is the effective rank metric defined in Equation~\ref{eq:er} 
\begin{equation}\label{eq:er}
    \mathrm{ER}(\theta_l) = \exp \left( - \sum _{k=1} ^ Q \frac{\sigma_k}{\norm{\boldsymbol{\sigma}}_1}  \log \frac{\sigma_k}{\norm{\boldsymbol{\sigma}}_1} \right)
\end{equation}
where $\boldsymbol{\sigma} =  \langle \sigma_1 \dots \sigma_Q \rangle$ is the vector of singular values of $\theta_l$ (in ascending order) and $\norm{\cdot}_1 $ is the $l_1$ norm. Stated differently, ER is the entropy computed over the normalised singular values of the weight matrix $\theta_l$. 

\begin{figure*}[t]
    \centering
    \includegraphics[width=\linewidth]{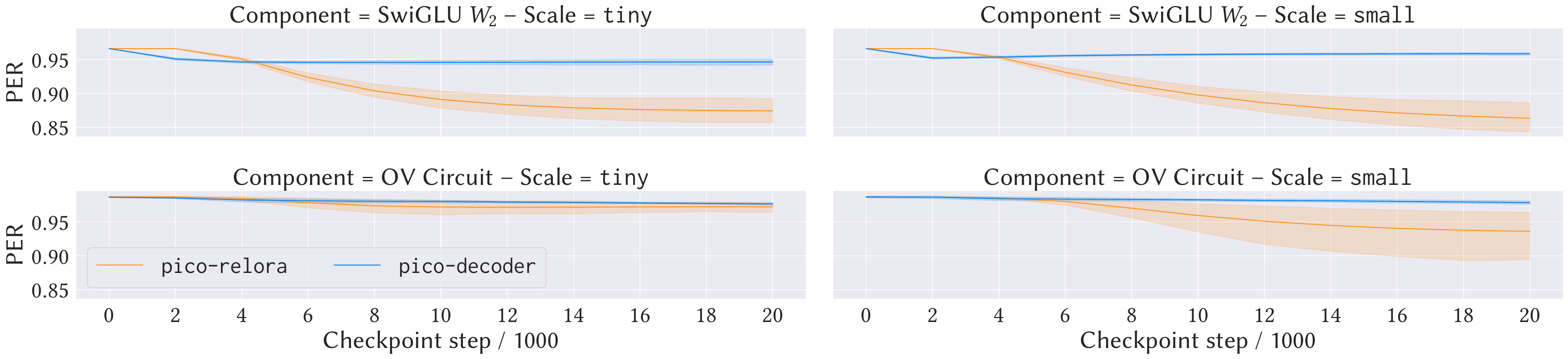}
    \caption{Proportional effective rank (PER) values of the parameters of the OV Circuit and the SwiGLU $W_2$ matrix, averaged over the models' layers. Values are shown with the 95\% confidence interval. }
    \label{fig:param-per}
\end{figure*}

\paragraph{PER of \texttt{pico-relora}'s weights declines throughout training.}Figure~\ref{fig:param-per} shows the layer-averaged PER of the OV Circuit and SwiGLU $W_2$ matrix during training, highlighting clear differences between \texttt{pico-relora} and \texttt{pico-decoder}. The former generally has a lower PER, with the gap widening as \texttt{pico-relora} steadily declines while \texttt{pico-decoder} remains flat; except for the \texttt{tiny} OV circuit, which shows little change and aligns closely with \texttt{pico-decoder}. Aside from this outlier, PER trends are consistent across scales. Confidence intervals for \texttt{pico-decoder} are consistently narrow, reflecting low layer-to-layer variability, whereas \texttt{pico-relora}'s widen over time.

\begin{figure*}[t]
    \centering
    \includegraphics[width=\linewidth]{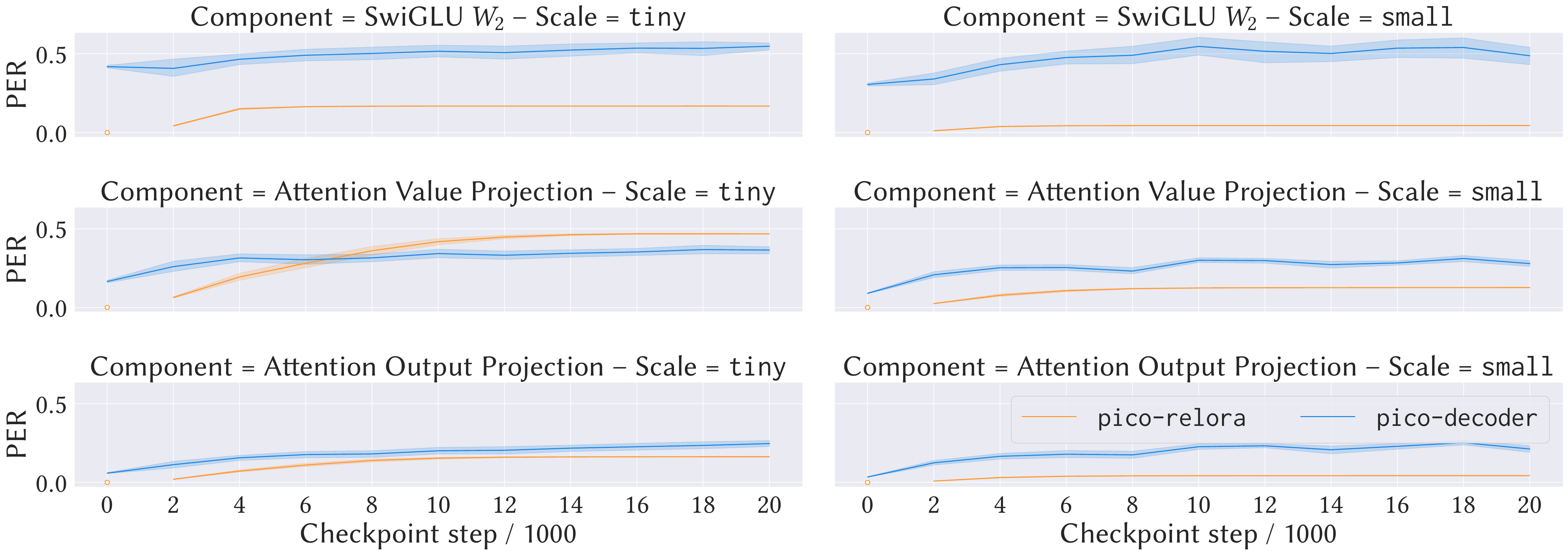}
    \caption{Proportional effective rank (PER) values of the gradient updates of the output and value projections in the attention mechanism and the SwiGLU $W_2$ matrix, averaged over the models' layers. Values are shown with bands representing the 95\% confidence interval. The hollow circles represent \texttt{NaN} values. }
    \label{fig:per-grad}
\end{figure*}

\paragraph{ReLoRA's gradient updates are rank deficient.}
Figure~\ref{fig:per-grad} shows PER on gradient updates for the Output and Value projections in attention and the SwiGLU $W_2$ projection. For \texttt{pico-decoder}, this is computed directly on gradients, while for \texttt{pico-relora}, the LoRA module weights serve as a proxy, since the frozen base matrices receive no gradients at each restart. Observations correspond to the weight measurements: the measured PER values are generally lower for ReLoRA, with this effect exacerbated at the \texttt{small} scale, with an outlier for attention at the \texttt{tiny} scale. 

\subsubsection{Condition number}

In addition to measured rank deficiencies, we explore further explanations for ReLoRA's underperformance. This leads us to investigate the model's susceptibility to error magnification, for which the condition number (CN or $\kappa$) of a matrix (introduced in Equation~\ref{eq:cond-num} below) can be a good proxy \citep{chapra_ebook_2011}. The CN measures how sensitive the solution of a linear system is to small changes in the input data or the matrix itself. Furthermore, a higher condition number can be used as a measure of the rounding errors arising from using finite precision arithmetic \citep[p.~208]{chapra_ebook_2011}, as is the case when training models. If the $\kappa$ of a matrix is $10^k$, then at least $k$ digits of precision are lost \citep[p.~321]{cheney_numerical_2008}. The condition number metric is simply the ratio of the largest to the smallest singular values of the input, which indicates how much the output can vary in response to slight variations in the input.

For the ordered vector of singular values $\boldsymbol{\sigma} = \langle \sigma_1 \dots \sigma_Q\rangle$ of some matrix $M$, the condition number $\kappa$ is thus
\begin{equation} \label{eq:cond-num}
    \kappa(M) = \frac{\sigma_Q}{\sigma_1}
\end{equation}

\begin{figure*}[t]
    \centering
    \includegraphics[width=\linewidth]{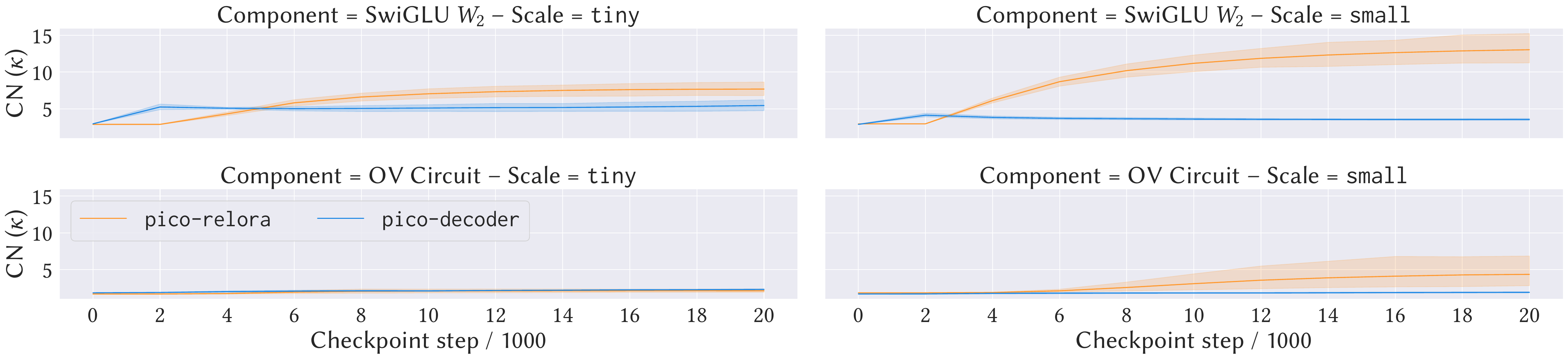}
    \caption{Condition numbers of the weight matrices of the OV Circuit and the SwiGLU $W_2$ matrix, averaged over the models' layers. Values are shown with the 95\% confidence interval. }
    \label{fig:param-cond-num}
\end{figure*}

\paragraph{\texttt{pico-relora} has larger CNs in the weight matrices.}
Figure~\ref{fig:param-cond-num} illustrates the condition numbers of the OV Circuit and SwiGLU $W_2$ matrices throughout training. The condition number is larger for \texttt{pico-relora} SwiGLU matrices, and even more on the \texttt{small} scale. For the OV circuit at the \texttt{tiny} scale, the $\kappa$ values are very similar, while at the \texttt{small} scale \texttt{pico-relora}'s values are greater with a wider confidence interval. 

\begin{figure*}[t]
    \centering
    \includegraphics[width=\linewidth]{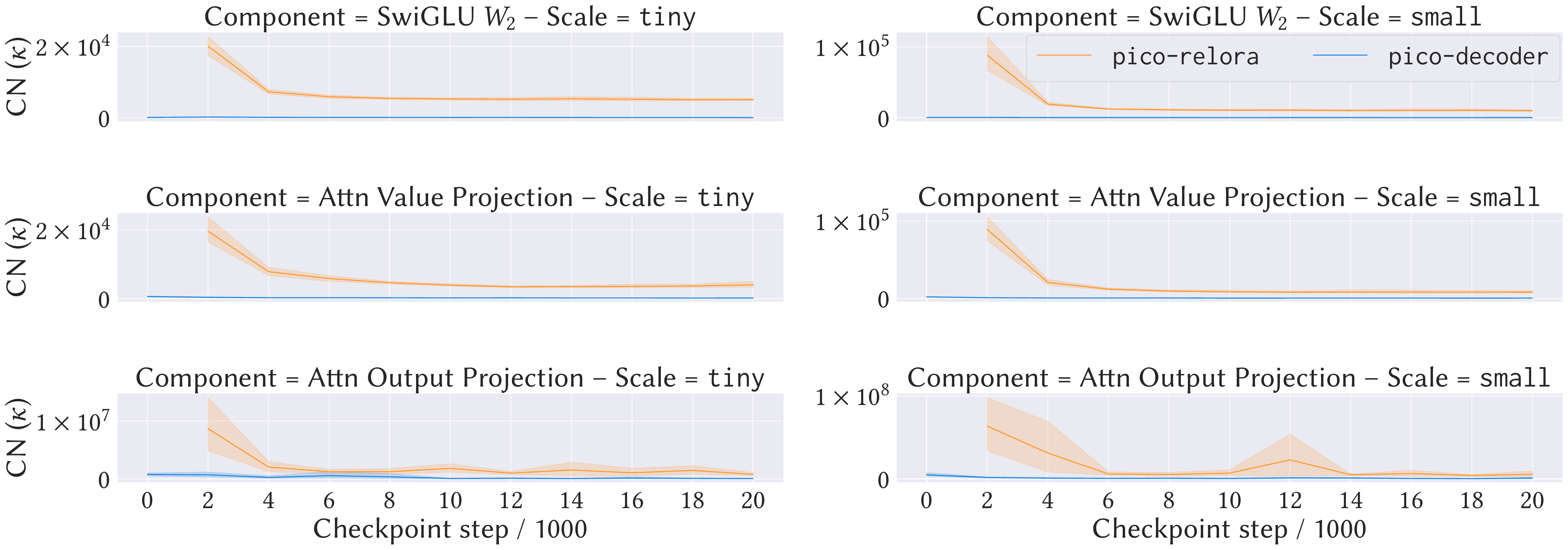}
        \caption{Condition numbers of the gradient updates of the output and value projections in the attention mechanism and the SwiGLU $W_2$ matrix, averaged over the models' layers. Values are shown with bands representing the 95\% confidence interval. The values at the initial checkpoints are omitted since they are zero matrices, with undefined $\kappa$. }
    \label{fig:cond-num-grad}
\end{figure*}

\paragraph{\texttt{pico-relora}'s gradient updates are \textit{highly ill-conditioned} early in training.}
Meanwhile, Figure~\ref{fig:cond-num-grad} shows the $\kappa$ values of the gradient updates of the output and value projections in the attention mechanism and the SwiGLU $W_2$ matrix. As can be seen, the condition numbers of the ReLoRA model's gradient updates initially start very high, indicating a highly ill-conditioned matrix, and then approach those of the baseline throughout training. These large $\kappa$ values suggest instability in the gradient updates and vastly increase round-off errors due to fixed precision, by up to eight additional digits of inaccuracy for \texttt{pico-relora-small}. For the same run's attention output projection, the condition number spikes at checkpoint step \num{12000}. 

\subsubsection{Findings from PER and CN}

\paragraph{\texttwemoji{thumbsdown} Low PER values \textit{drag} ReLoRA's performance down.}
Low PER values in the weight matrices and gradient updates exacerbate the rank deficiencies present in smaller models, thereby inhibiting performance compared to full-rank training.

\paragraph{\texttwemoji{thumbsdown} \texttt{pico-relora} is considerably more sensitive to fluctuations in inputs.}
The results for CN demonstrate that ReLoRA increases the sensitivity to error magnification caused by small variations in the input. Therefore, even tokens with relatively similar internal representations can have vastly different outputs. 

This behaviour is compounded by the \textit{anisotropy of SLMs}, which refers to the phenomenon where the internal representations of tokens are \textit{unevenly distributed} and \textit{highly clustered} within the representational space \citep{godey_anisotropy_2024}. As a result, even small changes in input can cause a token embedding to shift into a sparse or differently clustered region, thereby amplifying differences in the model's output. 

When both these effects combine, they lead to a model which is highly sensitive to any minor fluctuations in its inputs.   

\section{Conclusions}

We present the first systematic investigation of ReLoRA for SLM pretraining, evaluating models at 11M (\texttt{tiny}) and 66M (\texttt{small}) parameters across various efficiency and performance metrics. 

\subsection{Findings}

We identify three key findings:

\begin{enumerate}[label=\textbf{(\arabic*)}]

\item \textbf{\texttwemoji{thumbs down} Drag: ReLoRA degrades pretraining performance in SLMs.}
Across loss, Paloma perplexity, and BLiMP evaluation, ReLoRA consistently underperforms conventional full-rank training. The performance gap is minor for \texttt{tiny} models but grows substantially as model size increases to the \texttt{small} scale. However, this might be caused by minimal parameter differences between the two tiny models, leading to a surprising performance loss (compared to the findings of \cite{lialin_relora_2023} on large models) as model scale increases. One possible explanation is that the tiny model's parameterisation leaves little difference between ReLoRA and full-rank updates, therefore leading to a lower-than-expected drop in performance. 

\item \textbf{\texttwemoji{thumbs down} Drag: ReLoRA exacerbates low-rank bottlenecks and training instability.}
Learning dynamics analysis indicates that ReLoRA leads to reduced PERs of the models' parameters and gradient updates. Alongside this, ReLoRA induces highly ill-conditioned gradient updates that increase the models' susceptibility to numerical errors. Both of these exacerbate existing issues caused by the inherent anisotropy of SLM representations. 

\item \textbf{Parameter-efficient pretraining does not trivially extend from large to small models.}
Unlike in large LMs, where low-rank updates can still capture rich training signals, SLMs appear to have insufficient redundancy in their representations, resulting in greater sensitivity to ReLoRA's repeated low-rank projections.

\end{enumerate}

\subsection{Implications}

This paper highlights a principal limitation of low-rank pretraining for SLMs, suggesting that for a parameter-efficient training method to succeed in the low-compute domain, it likely must preserve higher-rank update spaces or adapt to models' intrinsic dimensionalities. Furthermore, our results prompt reflection on whether parameter-efficient training methods are necessary for SLMs at all. Unlike SOTA multi-billion-parameter models, SLMs can be trained on a single modern GPU (and often even on consumer hardware), raising the question of whether parameter-efficiency provides any benefits in this regime. 

While these are negative results, they are informative in highlighting the boundaries at which parameter-efficient pretraining breaks down. Our study thus highlights where new methods are most needed.

Future investigations should therefore focus on novel hybrid approaches. For instance, coupling low-rank adapters with selective full-rank updates or utilising dynamic rank adaptation techniques (such as DyLoRA \citep{valipour_dylora_2023}) to minimise representational losses. Our results aim to inform the design of future low-resource LM training algorithms, emphasising that efficiency gains cannot come at the cost of substantial expressivity and performance loss.

\section*{Limitations}

While our work offers valuable contributions, there are a few minor considerations to keep in mind:

\paragraph{(1) Single-run experiments.}
Due to constraints on computational resources, we performed only one training run per configuration. While the results are interesting, additional seeds at each scale would lead to more robust results. 

\paragraph{(2) Limited model scale and diversity.}
Likewise, the experiments were performed at only two scales and evaluated with a single decoder-only architecture. Investigating other intermediate scales and encoder-decoder models may provide further insight and more robust trends. 

\paragraph{(3) Limited hyperparameter exploration.}
ReLoRA involves several hyperparameters, such as restart frequency, LoRA rank, and optimiser state pruning ratio. Our study evaluates a single shared configuration and omits the full-rank warm start included by \citet{lialin_relora_2023}, which may not reflect the method's ideal performance.

\bibliography{latex/paper}

\begin{thebibliography}{37}
\providecommand{\natexlab}[1]{#1}

\bibitem[{Aghajanyan et~al.(2021)Aghajanyan, Gupta, and Zettlemoyer}]{aghajanyan_intrinsic_2021}
Armen Aghajanyan, Sonal Gupta, and Luke Zettlemoyer. 2021.
\newblock \href {https://doi.org/10.18653/v1/2021.acl-long.568} {Intrinsic {{Dimensionality Explains}} the {{Effectiveness}} of {{Language Model Fine-Tuning}}}.
\newblock In \emph{Proceedings of the 59th {{Annual Meeting}} of the {{Association}} for {{Computational Linguistics}} and the 11th {{International Joint Conference}} on {{Natural Language Processing}} ({{Volume}} 1: {{Long Papers}})}, pages 7319--7328, Online. Association for Computational Linguistics.

\bibitem[{Ainslie et~al.(2023)Ainslie, {Lee-Thorp}, {de Jong}, Zemlyanskiy, Lebron, and Sanghai}]{ainslie_gqa_2023}
Joshua Ainslie, James {Lee-Thorp}, Michiel {de Jong}, Yury Zemlyanskiy, Federico Lebron, and Sumit Sanghai. 2023.
\newblock \href {https://doi.org/10.18653/v1/2023.emnlp-main.298} {{{GQA}}: {{Training Generalized Multi-Query Transformer Models}} from {{Multi-Head Checkpoints}}}.
\newblock In \emph{Proceedings of the 2023 {{Conference}} on {{Empirical Methods}} in {{Natural Language Processing}}}, pages 4895--4901, Singapore. Association for Computational Linguistics.

\bibitem[{Blattmann et~al.(2023)Blattmann, Dockhorn, Kulal, Mendelevitch, Kilian, Lorenz, Levi, English, Voleti, Letts, Jampani, and Rombach}]{blattmann_stable_2023}
Andreas Blattmann, Tim Dockhorn, Sumith Kulal, Daniel Mendelevitch, Maciej Kilian, Dominik Lorenz, Yam Levi, Zion English, Vikram Voleti, Adam Letts, Varun Jampani, and Robin Rombach. 2023.
\newblock \href {https://doi.org/10.48550/arXiv.2311.15127} {Stable {{Video Diffusion}}: {{Scaling Latent Video Diffusion Models}} to {{Large Datasets}}}.
\newblock \emph{Preprint}, arXiv:2311.15127.

\bibitem[{{Boix-Adsera} et~al.(2023){Boix-Adsera}, Littwin, Abbe, Bengio, and Susskind}]{boix-adsera_transformers_2023}
Enric {Boix-Adsera}, Etai Littwin, Emmanuel Abbe, Samy Bengio, and Joshua Susskind. 2023.
\newblock \href {https://proceedings.neurips.cc/paper_files/paper/2023/hash/4d69c1c057a8bd570ba4a7b71aae8331-Abstract-Conference.html} {Transformers learn through gradual rank increase}.
\newblock \emph{Advances in Neural Information Processing Systems}, 36:24519--24551.

\bibitem[{Chapra(2011)}]{chapra_ebook_2011}
Steven Chapra. 2011.
\newblock \emph{{{EBOOK}}: {{Applied Numerical Methods}} with {{MATLAB}} for {{Engineers}} and {{Scientists}}}.
\newblock McGraw Hill.

\bibitem[{Cheney et~al.(2008)Cheney, Cheney, and Kincaid}]{cheney_numerical_2008}
Elliott~W. Cheney, Elliott~W. Cheney, and David Kincaid. 2008.
\newblock \emph{Numerical Mathematics and Computing}, 6. ed., internat. student ed edition.
\newblock Thomson Brooks/Cole, Belmont, Calif.

\bibitem[{Chowdhery et~al.(2023)Chowdhery, Narang, Devlin, Bosma, Mishra, Roberts, Barham, Chung, Sutton, Gehrmann, Schuh, Shi, Tsvyashchenko, Maynez, Rao, Barnes, Tay, Shazeer, Prabhakaran, Reif, Du, Hutchinson, Pope, Bradbury, Austin, Isard, {Gur-Ari}, Yin, Duke, Levskaya, Ghemawat, Dev, Michalewski, Garcia, Misra, Robinson, Fedus, Zhou, Ippolito, Luan, Lim, Zoph, Spiridonov, Sepassi, Dohan, Agrawal, Omernick, Dai, Pillai, Pellat, Lewkowycz, Moreira, Child, Polozov, Lee, Zhou, Wang, Saeta, Diaz, Firat, Catasta, Wei, {Meier-Hellstern}, Eck, Dean, Petrov, and Fiedel}]{chowdhery_palm_2023}
Aakanksha Chowdhery, Sharan Narang, Jacob Devlin, Maarten Bosma, Gaurav Mishra, Adam Roberts, Paul Barham, Hyung~Won Chung, Charles Sutton, Sebastian Gehrmann, Parker Schuh, Kensen Shi, Sasha Tsvyashchenko, Joshua Maynez, Abhishek Rao, Parker Barnes, Yi~Tay, Noam Shazeer, Vinodkumar Prabhakaran, and 48 others. 2023.
\newblock \href {http://jmlr.org/papers/v24/22-1144.html} {{{PaLM}}: {{Scaling}} language modeling with pathways}.
\newblock \emph{Journal of Machine Learning Research}, 24(240):1--113.

\bibitem[{Dao(2023)}]{dao_flashattention-2_2023}
Tri Dao. 2023.
\newblock \href {https://doi.org/10.48550/arXiv.2307.08691} {{{FlashAttention-2}}: {{Faster Attention}} with {{Better Parallelism}} and {{Work Partitioning}}}.
\newblock \emph{Preprint}, arXiv:2307.08691.

\bibitem[{Dettmers et~al.(2023)Dettmers, Pagnoni, Holtzman, and Zettlemoyer}]{dettmers_qlora_2023}
Tim Dettmers, Artidoro Pagnoni, Ari Holtzman, and Luke Zettlemoyer. 2023.
\newblock \href {https://proceedings.neurips.cc/paper_files/paper/2023/hash/1feb87871436031bdc0f2beaa62a049b-Abstract-Conference.html} {{{QLoRA}}: {{Efficient Finetuning}} of {{Quantized LLMs}}}.
\newblock \emph{Advances in Neural Information Processing Systems}, 36:10088--10115.

\bibitem[{Diehl~Martinez(2025)}]{diehl_martinez_pico_2025}
Richard Diehl~Martinez. 2025.
\newblock \href {https://github.com/pico-lm} {Pico: A lightweight framework for studying language model learning dynamics}.

\bibitem[{Diehl~Martinez et~al.(2024)Diehl~Martinez, Lesci, and Buttery}]{diehl_martinez_tending_2024}
Richard Diehl~Martinez, Pietro Lesci, and Paula Buttery. 2024.
\newblock \href {https://aclanthology.org/2024.findings-emnlp.187} {Tending {{Towards Stability}}: {{Convergence Challenges}} in {{Small Language Models}}}.
\newblock In \emph{Findings of the {{Association}} for {{Computational Linguistics}}: {{EMNLP}} 2024}, pages 3275--3286, Miami, Florida, USA. Association for Computational Linguistics.

\bibitem[{Elhage et~al.(2021)Elhage, Nanda, Olsson, Henighan, Joseph, Mann, Askell, Bai, Chen, Conerly et~al.}]{elhage_mathematical_2021}
Nelson Elhage, Neel Nanda, Catherine Olsson, Tom Henighan, Nicholas Joseph, Ben Mann, Amanda Askell, Yuntao Bai, Anna Chen, Tom Conerly, and 1 others. 2021.
\newblock A mathematical framework for transformer circuits.
\newblock \emph{Transformer Circuits Thread}, 1(1):12.

\bibitem[{Ettinger(2020)}]{ettinger_what_2020}
Allyson Ettinger. 2020.
\newblock \href {https://doi.org/10.1162/tacl_a_00298} {What {{BERT Is Not}}: {{Lessons}} from a {{New Suite}} of {{Psycholinguistic Diagnostics}} for {{Language Models}}}.
\newblock \emph{Transactions of the Association for Computational Linguistics}, 8:34--48.

\bibitem[{Fomenko et~al.(2024)Fomenko, Yu, Lee, Hsieh, and Chen}]{fomenko_note_2024}
Vlad Fomenko, Han Yu, Jongho Lee, Stanley Hsieh, and Weizhu Chen. 2024.
\newblock \href {https://doi.org/10.48550/arXiv.2404.05086} {A {{Note}} on {{LoRA}}}.
\newblock \emph{Preprint}, arXiv:2404.05086.

\bibitem[{Godey et~al.(2024{\natexlab{a}})Godey, Clergerie, and Sagot}]{godey_anisotropy_2024}
Nathan Godey, {\'E}ric Clergerie, and Beno{\^i}t Sagot. 2024{\natexlab{a}}.
\newblock \href {https://aclanthology.org/2024.eacl-long.3/} {Anisotropy {{Is Inherent}} to {{Self-Attention}} in {{Transformers}}}.
\newblock In \emph{Proceedings of the 18th {{Conference}} of the {{European Chapter}} of the {{Association}} for {{Computational Linguistics}} ({{Volume}} 1: {{Long Papers}})}, pages 35--48, St. Julian's, Malta. Association for Computational Linguistics.

\bibitem[{Godey et~al.(2024{\natexlab{b}})Godey, de~la Clergerie, and Sagot}]{godey_why_2024}
Nathan Godey, {\'E}ric~Villemonte de~la Clergerie, and Beno{\^i}t Sagot. 2024{\natexlab{b}}.
\newblock \href {https://openreview.net/forum?id=MoitXWlXcS#discussion} {Why do small language models underperform? {{Studying Language Model Saturation}} via the {{Softmax Bottleneck}}}.
\newblock In \emph{First {{Conference}} on {{Language Modeling}}}.

\bibitem[{Grattafiori et~al.(2024)Grattafiori, Dubey, Jauhri, Pandey, Kadian, {Al-Dahle}, Letman, Mathur, Schelten, Vaughan et~al.}]{grattafiori_llama_2024}
Aaron Grattafiori, Abhimanyu Dubey, Abhinav Jauhri, Abhinav Pandey, Abhishek Kadian, Ahmad {Al-Dahle}, Aiesha Letman, Akhil Mathur, Alan Schelten, Alex Vaughan, and 1 others. 2024.
\newblock \href {https://arxiv.org/abs/2407.21783} {The llama 3 herd of models}.
\newblock \emph{arXiv preprint arXiv:2407.21783}.

\bibitem[{Groeneveld et~al.(2024)Groeneveld, Beltagy, Walsh, Bhagia, Kinney, Tafjord, Jha, Ivison, Magnusson, Wang, Arora, Atkinson, Authur, Chandu, Cohan, Dumas, Elazar, Gu, Hessel, Khot, Merrill, Morrison, Muennighoff, Naik, Nam, Peters, Pyatkin, Ravichander, Schwenk, Shah, Smith, Strubell, Subramani, Wortsman, Dasigi, Lambert, Richardson, Zettlemoyer, Dodge, Lo, Soldaini, Smith, and Hajishirzi}]{groeneveld_olmo_2024}
Dirk Groeneveld, Iz~Beltagy, Evan Walsh, Akshita Bhagia, Rodney Kinney, Oyvind Tafjord, Ananya Jha, Hamish Ivison, Ian Magnusson, Yizhong Wang, Shane Arora, David Atkinson, Russell Authur, Khyathi Chandu, Arman Cohan, Jennifer Dumas, Yanai Elazar, Yuling Gu, Jack Hessel, and 24 others. 2024.
\newblock \href {https://doi.org/10.18653/v1/2024.acl-long.841} {{{OLMo}}: {{Accelerating}} the {{Science}} of {{Language Models}}}.
\newblock In \emph{Proceedings of the 62nd {{Annual Meeting}} of the {{Association}} for {{Computational Linguistics}} ({{Volume}} 1: {{Long Papers}})}, pages 15789--15809, Bangkok, Thailand. Association for Computational Linguistics.

\bibitem[{Hu et~al.(2022)Hu, Shen, Wallis, {Allen-Zhu}, Li, Wang, Wang, and Chen}]{hu_lora_2022}
Edward~J Hu, Yelong Shen, Phillip Wallis, Zeyuan {Allen-Zhu}, Yuanzhi Li, Shean Wang, Lu~Wang, and Weizhu Chen. 2022.
\newblock \href {https://openreview.net/forum?id=nZeVKeeFYf9} {{{LoRA}}: {{Low-rank}} adaptation of large language models}.
\newblock In \emph{International Conference on Learning Representations}.

\bibitem[{Kaplan et~al.(2020)Kaplan, McCandlish, Henighan, Brown, Chess, Child, Gray, Radford, Wu, and Amodei}]{kaplan_scaling_2020}
Jared Kaplan, Sam McCandlish, Tom Henighan, Tom~B. Brown, Benjamin Chess, Rewon Child, Scott Gray, Alec Radford, Jeffrey Wu, and Dario Amodei. 2020.
\newblock \href {https://doi.org/10.48550/arXiv.2001.08361} {Scaling {{Laws}} for {{Neural Language Models}}}.
\newblock \emph{Preprint}, arXiv:2001.08361.

\bibitem[{Lialin et~al.(2023)Lialin, Muckatira, Shivagunde, and Rumshisky}]{lialin_relora_2023}
Vladislav Lialin, Sherin Muckatira, Namrata Shivagunde, and Anna Rumshisky. 2023.
\newblock \href {https://openreview.net/forum?id=DLJznSp6X3} {{{ReLoRA}}: {{High-Rank Training Through Low-Rank Updates}}}.
\newblock In \emph{The {{Twelfth International Conference}} on {{Learning Representations}}}.

\bibitem[{Liu et~al.(2019)Liu, Ott, Goyal, Du, Joshi, Chen, Levy, Lewis, Zettlemoyer, and Stoyanov}]{liu_roberta_2019-2}
Yinhan Liu, Myle Ott, Naman Goyal, Jingfei Du, Mandar Joshi, Danqi Chen, Omer Levy, Mike Lewis, Luke Zettlemoyer, and Veselin Stoyanov. 2019.
\newblock \href {https://doi.org/10.48550/arXiv.1907.11692} {{{RoBERTa}}: {{A Robustly Optimized BERT Pretraining Approach}}}.
\newblock \emph{Preprint}, arXiv:1907.11692.

\bibitem[{Loshchilov and Hutter(2019)}]{loshchilov_decoupled_2019}
Ilya Loshchilov and Frank Hutter. 2019.
\newblock \href {https://doi.org/10.48550/arXiv.1711.05101} {Decoupled {{Weight Decay Regularization}}}.
\newblock \emph{Preprint}, arXiv:1711.05101.

\bibitem[{Magnusson et~al.(2024)Magnusson, Bhagia, Hofmann, Soldaini, Jha, Tafjord, Schwenk, Walsh, Elazar, Lo et~al.}]{magnusson_paloma_2024}
Ian Magnusson, Akshita Bhagia, Valentin Hofmann, Luca Soldaini, Ananya~Harsh Jha, Oyvind Tafjord, Dustin Schwenk, Evan Walsh, Yanai Elazar, Kyle Lo, and 1 others. 2024.
\newblock Paloma: {{A}} benchmark for evaluating language model fit.
\newblock \emph{Advances in Neural Information Processing Systems}, 37:64338--64376.

\bibitem[{Mielke et~al.(2019)Mielke, Cotterell, Gorman, Roark, and Eisner}]{mielke_what_2019}
Sebastian~J. Mielke, Ryan Cotterell, Kyle Gorman, Brian Roark, and Jason Eisner. 2019.
\newblock \href {https://doi.org/10.18653/v1/P19-1491} {What {{Kind}} of {{Language Is Hard}} to {{Language-Model}}?}
\newblock In \emph{Proceedings of the 57th {{Annual Meeting}} of the {{Association}} for {{Computational Linguistics}}}, pages 4975--4989, Florence, Italy. Association for Computational Linguistics.

\bibitem[{Morrison et~al.(2024)Morrison, Na, Fernandez, Dettmers, Strubell, and Dodge}]{morrison_holistically_2024}
Jacob Morrison, Clara Na, Jared Fernandez, Tim Dettmers, Emma Strubell, and Jesse Dodge. 2024.
\newblock \href {https://openreview.net/forum?id=04qx93Viwj} {Holistically {{Evaluating}} the {{Environmental Impact}} of {{Creating Language Models}}}.
\newblock In \emph{The {{Thirteenth International Conference}} on {{Learning Representations}}}.

\bibitem[{Noci et~al.(2022)Noci, Anagnostidis, Biggio, Orvieto, Singh, and Lucchi}]{noci_signal_2022}
Lorenzo Noci, Sotiris Anagnostidis, Luca Biggio, Antonio Orvieto, Sidak~Pal Singh, and Aurelien Lucchi. 2022.
\newblock \href {https://proceedings.neurips.cc/paper_files/paper/2022/hash/ae0cba715b60c4052359b3d52a2cff7f-Abstract-Conference.html} {Signal {{Propagation}} in {{Transformers}}: {{Theoretical Perspectives}} and the {{Role}} of {{Rank Collapse}}}.
\newblock \emph{Advances in Neural Information Processing Systems}, 35:27198--27211.

\bibitem[{{OpenAI}(2024)}]{openai_gpt-4o_2024}
{OpenAI}. 2024.
\newblock \href {https://doi.org/10.48550/arXiv.2410.21276} {{{GPT-4o System Card}}}.
\newblock \emph{Preprint}, arXiv:2410.21276.

\bibitem[{Roy and Vetterli(2007)}]{roy_effective_2007}
Olivier Roy and Martin Vetterli. 2007.
\newblock \href {https://ieeexplore.ieee.org/document/7098875/?arnumber=7098875} {The effective rank: {{A}} measure of effective dimensionality}.
\newblock In \emph{2007 15th {{European Signal Processing Conference}}}, pages 606--610.

\bibitem[{Schwartz et~al.(2020)Schwartz, Dodge, Smith, and Etzioni}]{schwartz_green_2020}
Roy Schwartz, Jesse Dodge, Noah~A. Smith, and Oren Etzioni. 2020.
\newblock \href {https://doi.org/10.1145/3381831} {Green {{AI}}}.
\newblock \emph{Communications of the ACM}, 63(12):54--63.

\bibitem[{Shazeer(2020)}]{shazeer_glu_2020}
Noam Shazeer. 2020.
\newblock \href {https://doi.org/10.48550/arXiv.2002.05202} {{{GLU Variants Improve Transformer}}}.
\newblock \emph{Preprint}, arXiv:2002.05202.

\bibitem[{Soldaini et~al.(2024)Soldaini, Kinney, Bhagia, Schwenk, Atkinson, Authur, Bogin, Chandu, Dumas, Elazar, Hofmann, Jha, Kumar, Lucy, Lyu, Lambert, Magnusson, Morrison, Muennighoff, Naik, Nam, Peters, Ravichander, Richardson, Shen, Strubell, Subramani, Tafjord, Walsh, Zettlemoyer, Smith, Hajishirzi, Beltagy, Groeneveld, Dodge, and Lo}]{soldaini_dolma_2024}
Luca Soldaini, Rodney Kinney, Akshita Bhagia, Dustin Schwenk, David Atkinson, Russell Authur, Ben Bogin, Khyathi Chandu, Jennifer Dumas, Yanai Elazar, Valentin Hofmann, Ananya Jha, Sachin Kumar, Li~Lucy, Xinxi Lyu, Nathan Lambert, Ian Magnusson, Jacob Morrison, Niklas Muennighoff, and 17 others. 2024.
\newblock \href {https://doi.org/10.18653/v1/2024.acl-long.840} {Dolma: An {{Open Corpus}} of {{Three Trillion Tokens}} for {{Language Model Pretraining Research}}}.
\newblock In \emph{Proceedings of the 62nd {{Annual Meeting}} of the {{Association}} for {{Computational Linguistics}} ({{Volume}} 1: {{Long Papers}})}, pages 15725--15788, Bangkok, Thailand. Association for Computational Linguistics.

\bibitem[{Su et~al.(2023)Su, Lu, Pan, Murtadha, Wen, and Liu}]{su_roformer_2023}
Jianlin Su, Yu~Lu, Shengfeng Pan, Ahmed Murtadha, Bo~Wen, and Yunfeng Liu. 2023.
\newblock \href {https://doi.org/10.48550/arXiv.2104.09864} {{{RoFormer}}: {{Enhanced Transformer}} with {{Rotary Position Embedding}}}.
\newblock \emph{Preprint}, arXiv:2104.09864.

\bibitem[{Touvron et~al.(2023)Touvron, Lavril, Izacard, Martinet, Lachaux, Lacroix, Rozi{\`e}re, Goyal, Hambro, Azhar et~al.}]{touvron_llama_2023}
Hugo Touvron, Thibaut Lavril, Gautier Izacard, Xavier Martinet, Marie-Anne Lachaux, Timoth{\'e}e Lacroix, Baptiste Rozi{\`e}re, Naman Goyal, Eric Hambro, Faisal Azhar, and 1 others. 2023.
\newblock \href {https://arxiv.org/abs/2302.13971} {Llama: {{Open}} and efficient foundation language models}.
\newblock \emph{arXiv preprint arXiv:2302.13971}.

\bibitem[{Valipour et~al.(2023)Valipour, Rezagholizadeh, Kobyzev, and Ghodsi}]{valipour_dylora_2023}
Mojtaba Valipour, Mehdi Rezagholizadeh, Ivan Kobyzev, and Ali Ghodsi. 2023.
\newblock \href {https://doi.org/10.18653/v1/2023.eacl-main.239} {{{DyLoRA}}: {{Parameter-Efficient Tuning}} of {{Pre-trained Models}} using {{Dynamic Search-Free Low-Rank Adaptation}}}.
\newblock In \emph{Proceedings of the 17th {{Conference}} of the {{European Chapter}} of the {{Association}} for {{Computational Linguistics}}}, pages 3274--3287, Dubrovnik, Croatia. Association for Computational Linguistics.

\bibitem[{Warstadt et~al.(2020)Warstadt, Parrish, Liu, Mohananey, Peng, Wang, and Bowman}]{warstadt_blimp_2020}
Alex Warstadt, Alicia Parrish, Haokun Liu, Anhad Mohananey, Wei Peng, Sheng-Fu Wang, and Samuel~R. Bowman. 2020.
\newblock \href {https://doi.org/10.1162/tacl_a_00321} {{{BLiMP}}: {{The Benchmark}} of {{Linguistic Minimal Pairs}} for {{English}}}.
\newblock \emph{Transactions of the Association for Computational Linguistics}, 8:377--392.

\bibitem[{Zhang and Sennrich(2019)}]{zhang_root_2019}
Biao Zhang and Rico Sennrich. 2019.
\newblock \href {https://proceedings.neurips.cc/paper/2019/hash/1e8a19426224ca89e83cef47f1e7f53b-Abstract.html} {Root {{Mean Square Layer Normalization}}}.
\newblock In \emph{Advances in {{Neural Information Processing Systems}}}, volume~32. Curran Associates, Inc.

\end{thebibliography}

\appendix

\section{The \texttt{pico-decoder} model} \label{sec:pico-decoder}

\begin{figure}
    \centering
    \includegraphics[width=0.9\linewidth]{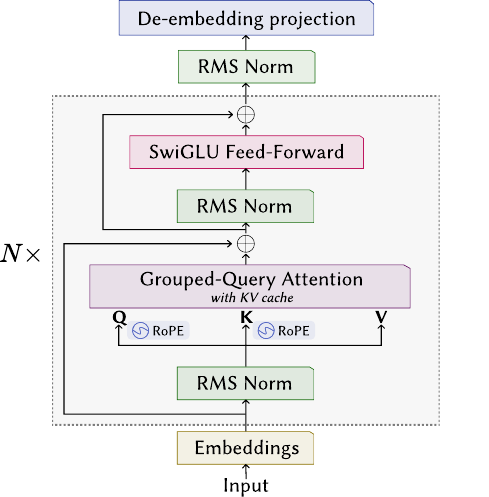}
    \caption{\texttt{pico-decoder}, a Llama-based model}
    \label{fig:pico-decoder}
\end{figure}

\texttt{pico-decoder} \citep{diehl_martinez_pico_2025} is a Llama-based \citep{touvron_llama_2023} model, which forms the base of the ablation study investigated in this work. It is illustrated in Figure~\ref{fig:pico-decoder}.

\texttt{pico-decoder} includes the following features and optimisations: 
\begin{itemize}
    \item Root-Mean-Square Layer Normalisation \citep{zhang_root_2019}
    \item RoPE embeddings \citep{su_roformer_2023}
    \item Grouped-query attention \citep{ainslie_gqa_2023} with key-value caching
    \item FlashAttention-2 \citep{dao_flashattention-2_2023}
    \item SwiGLU activations \citep{shazeer_glu_2020}
\end{itemize}

\section{Training infrastructure and runtimes}
\label{sec:infrastructure}

The \texttt{pico-relora} training runs were conducted on a single Ampere GPU node (four GPUs) each, hosted by CSD3. By contrast, the \texttt{pico-decoder} runs were performed on four Ampere nodes (sixteen GPUs). 

The Ampere nodes are Dell PowerEdge XE8545 servers consisting of two AMD EPYC 7763 64-Core Processors \qty{1.8}{\giga\hertz}, \qty{1000}{\gibi\byte} RAM, four NVIDIA A100-SXM-80GB GPUs, and dual-rail Mellanox HDR200 InfiniBand interconnect. Each A100 GPU is made up of \num{6912} FP32 CUDA Cores.\footnote{This information is sourced here: \url{https://docs.hpc.cam.ac.uk/hpc/user-guide/a100.html}.}

The wall-clock runtime and total GPU hours of each of the runs are shown in Table~\ref{tab:runtimes}. For the \texttt{tiny} models, \texttt{pico-relora} takes 13.5\% fewer GPU hours to train, while for the \texttt{small} scale, it was 14.7\% more efficient. These are not entirely equivalent comparisons, however, as the \texttt{pico-relora} runs were configured to checkpoint and evaluate more frequently and with more gradient accumulation steps, alongside other minor differences in the runtime environment. 

\begin{table}[t]
    \centering
    \begin{tabular}{cS[table-format=3.2]S[table-format=3.2]}
    \toprule
    \textbf{Model} & {\textbf{Wall-clock time} / h} & \textbf{GPU hours}   \\
    \midrule
    \texttt{t-dec} & 52.96 & 847.37 \\
    \texttt{s-dec} & 61.29 & 980.56 \\ 
    \texttt{t-rel} & 183.17 & 732.70 \\
    \texttt{s-rel} & 215.87 & 863.47 \\
    \bottomrule
    \end{tabular}
    \caption{Wallclock time and equivalent GPU hours for each of the four training runs. \texttt{t} = \texttt{tiny}, \texttt{s} = \texttt{small}, \texttt{dec} = \texttt{decoder}, \texttt{rel} = \texttt{relora}. }
    \label{tab:runtimes}
\end{table}

\section{Full hyperparameter configurations} \label{app:full_conf}

\subsection{Learning rate scheduler and optimiser configuration -- {\itshape shared}}

\begin{figure}
    \centering
    \includegraphics[width=\linewidth]{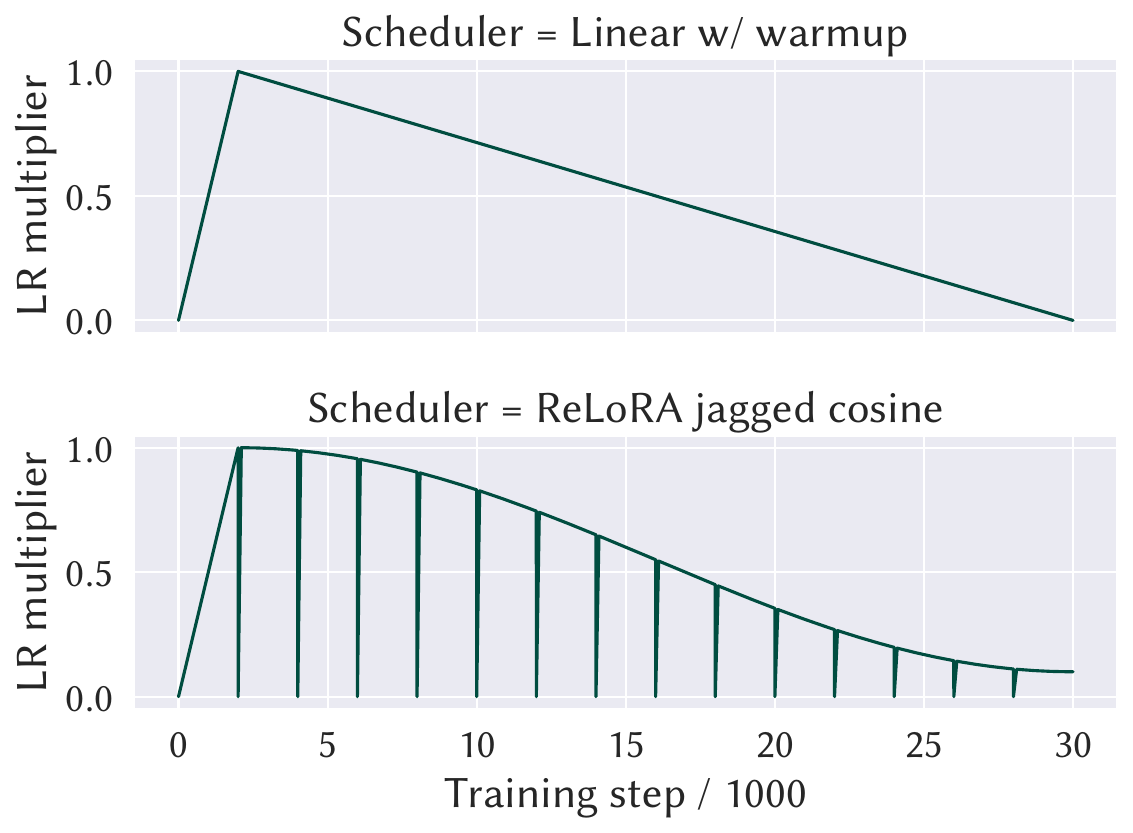}
    \caption{Graph showing the linear LR schedule used by \texttt{pico-decoder} and ReLoRA's jagged cosine scheduler \citep{lialin_relora_2023}. Both are shown here, configured with a total of \num{30000} steps and an initial warmup of \num{2000} steps. The ReLoRA scheduler is additionally configured with \num{100} post-restart warmup steps and a reset frequency of \num{2000}.}
    \label{fig:lr-schedulers}
\end{figure}

Table~\ref{tab:lr_opt_config} depicts the learning rate and optimisation configurations for both runs. It has an initial linear warmup of \num{2000} steps, and resets in line with the ReLoRA frequency specified in Table~\ref{tab:relora_config}. After each reset, the learning rate is linearly re-warmed over \num{100} steps before the standard cosine decay is resumed. The learning rate decays down to a minimum of $10\%$ of the initial rate. The baseline runs use the linear scheduler configured with a \num{2500} step initial warmup. However, as this is indivisible by the \num{2000} step ReLoRA frequency, the jagged cosine scheduler uses the nearest multiple, a \num{2000} step initial warmup. Both schedulers are shown in Figure~\ref{fig:lr-schedulers}.

\begin{table}[H]
    \centering
    \begin{tabular}{lr}
    \toprule
      \textbf{Parameter} & \textbf{Value} \\
    \midrule
       Optimiser  & \texttt{adamw}  \\
       Learning rate & \num{3e-4} \\ 
       LR scheduler & \verb|relora_jagged_cosine| \\
       LR warmup steps & \num{2000}  \\
       Min LR ratio & \num{0.1} \\
       Restart warmup steps & \num{100}\\
    \bottomrule 
    \end{tabular}
    \caption{Learning rate scheduler and optimiser configuration for training runs}
    \label{tab:lr_opt_config}
\end{table}

\subsection{Data configuration -- {\itshape shared}}

Table~\ref{tab:data_config} configures the data pipeline for the training runs. The training and perplexity datasets are \texttt{pretokenized-dolma} and \texttt{pretokenized-paloma-tinsy}, as described in Section~\ref{sec:impl-datasets}. The tokeniser used, \texttt{allenai/OLMo-7B-0724-hf}, is the one used for the OLMo model\footnote{Which can be found on HuggingFace here: \url{https://huggingface.co/allenai/OLMo-7B-0724-hf}.} \citep{groeneveld_olmo_2024}.

\begin{table}[H]
    \centering
    \begin{tabular}{lr}
    \toprule
      \textbf{Parameter} & \textbf{Value} \\
    \midrule
    Vocab size  & \num{50304}  \\
    Batch size & \num{1024} \\ 
    Max seq length & \num{2048}  \\
    \bottomrule 
    \end{tabular}
    \caption{Dataset configuration}
    \label{tab:data_config}
\end{table}

\subsection{Model configuration -- {\itshape shared}}

Table~\ref{tab:model_config} depicts the model configuration parameters for the \texttt{tiny} and \texttt{small} training runs. The two runs differ only in their $d_\mathrm{model}$ and $d_\mathrm{ff}$ parameters, keeping the depth of the model and attention architecture constant.

\begin{table}[H]
\centering
    \begin{tabular}{lcc}
    \toprule
      \textbf{Parameter} & \textbf{\texttt{tiny}} &\textbf{\texttt{small}} \\
    \midrule
       $n_\mathrm{layers}$ & \multicolumn{2}{c}{\num{12}} \\
       $n_\mathrm{heads}$ & \multicolumn{2}{c}{\num{12}} \\
       $n_\mathrm{heads_{kv}}$ & \multicolumn{2}{c}{\num{4}} \\ 
       $d_\mathrm{model}$ & $96$ & $384$ \\
       $d_\mathrm{ff}$ & $384$ & $1536$ \\
    \bottomrule 
    \end{tabular}
    \caption{Model configuration for training runs}
    \label{tab:model_config}
\end{table}

\end{document}